\documentclass[10pt,journal,compsoc]{IEEEtran}

\ifCLASSOPTIONcompsoc
  \usepackage[nocompress]{cite}
\else
  \usepackage{cite}
\fi

\ifCLASSINFOpdf
\else
\fi

\usepackage{algorithmic}

\usepackage{graphicx}
\graphicspath{{Figures/}}
\usepackage{amsmath,amssymb}
\usepackage{textcomp,upgreek,gensymb}
\usepackage{enumitem}
\setitemize{noitemsep,topsep=0pt,parsep=0pt,partopsep=0pt}
\usepackage{booktabs}
\usepackage{multirow}
\usepackage{subcaption}
\usepackage{color}
\usepackage[figuresright]{rotating}
\usepackage{ragged2e}
\usepackage{fontawesome}
\usepackage{caption}
\usepackage[pagebackref=true,breaklinks=true,letterpaper=true,colorlinks,bookmarks=false,urlcolor=red,citecolor=cyan]{hyperref}

\begin{document}
\title{\raisebox{-0.5em}{\includegraphics[height=1.8em]{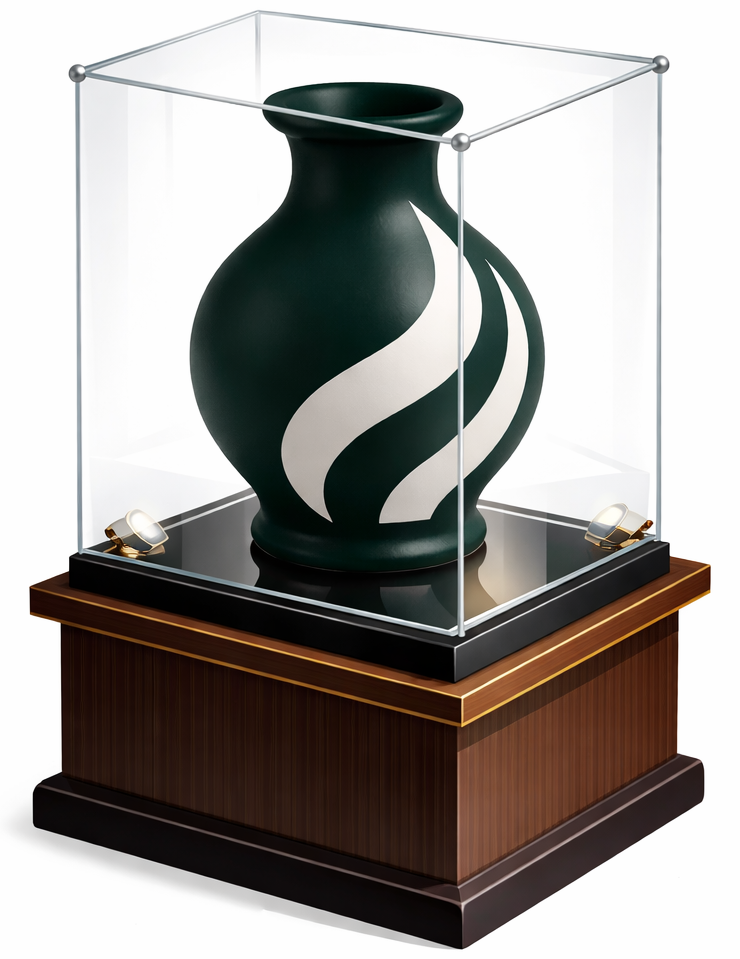}}~VaseMuseum: Digital Intelligent Museum for Ancient Greek Pottery}%

\author{Jiazi Wang$^*$,
    Nonghai Zhang$^*$,
    Qiushi Xie$^*$,
    Zeyu Zhang$^{*\dag}$,
    Yufeng Chen,
    Yang Zhao,\\
    Ling Shao,~\IEEEmembership{Fellow,~IEEE,}
    Hao Tang$^\ddag$
	\IEEEcompsocitemizethanks{
        \IEEEcompsocthanksitem $^*$Equal contribution. $^\dag$Project lead. \protect
        \IEEEcompsocthanksitem $^\ddag$Corresponding author, E-mail: bjdxtanghao@gmail.com. \protect
        \IEEEcompsocthanksitem Jiazi Wang and Yufeng Chen are with School of Computer Science and Technology, Beijing Jiaotong University, Beijing 100044, China. \protect
        \IEEEcompsocthanksitem Nonghai Zhang, Zeyu Zhang, and Hao Tang are with School of Computer Science, Peking University, Beijing 100871, China. \protect
        \IEEEcompsocthanksitem Qiushi Xie is with Huazhong University of Science and Technology, Wuhan 430074, China.
        \IEEEcompsocthanksitem Yang Zhao is with the Department of Computer Science and Information Technology, La Trobe University, Melbourne VIC 3000, Australia. \protect
        \IEEEcompsocthanksitem  Ling Shao is with the UCAS-Terminus AI Lab, University of Chinese Academy of Sciences, Beijing 100049, China 
        }%
}

\markboth{Submitted to IEEE Transactions on Pattern Analysis and Machine Intelligence}%
{Wang \MakeLowercase{\textit{et al.}}: VaseMuseum: Digital Intelligent Museum for Ancient Greek Pottery}

\IEEEtitleabstractindextext{%
\justify
\begin{abstract}
Vision-language models (VLMs) have made interactive digital museums increasingly feasible by connecting 3D digitization with natural-language artifact exploration.
However, in cultural heritage domains such as ancient Greek pottery, reliable VLM assistance is limited by two challenges.
First, open-ended interpretation requires grounding fine-grained 2D/3D visual evidence in specialized curatorial knowledge, yet the retrieval process may introduce weak sources and unverifiable references.
Second, when the available evidence is incomplete, noisy, or ambiguous, VLMs often produce confident but unsupported answers instead of calibrated uncertainty.
To address these challenges, we propose VaseMuseum, a lightweight and modular multimodal agent framework for intelligent digital museums of ancient Greek pottery.
VaseMuseum combines an interactive virtual museum with VaseAgent, which supports both 2D images and 3D artifacts through multimodal perception, 3D-aware reasoning, external knowledge retrieval, and inference-time reliability control. 
Specifically, VaseAgent retrieves evidence from authoritative web and museum knowledge sources, and source-level control selects diverse and verifiable evidence before generation. 
Meanwhile, response-level control checks generated claims against the evidence pool and encourages neutral, evidence-bounded answers when support is insufficient or conflicting. 
Moreover, a training-free GRPO-style selection mechanism favors responses with valid references and calibrated confidence without updating the VLM backbone. Experiments in a realistic digital museum simulation show that VaseMuseum improves citation validity, reduces hallucinations on knowledge-intensive queries, and produces more neutral answers under ambiguity compared with search-enabled VLM baselines. These results suggest a practical path toward trustworthy multimodal systems for cultural heritage applications.
Code:~\url{https://github.com/AIGeeksGroup/VaseMuseum}.
Website:~\url{https://aigeeksgroup.github.io/VaseMuseum}.

\end{abstract}

\begin{IEEEkeywords}
Large language models, multimodal learning, vision-language models, visual question answering.
\end{IEEEkeywords}}

\maketitle

\IEEEdisplaynontitleabstractindextext

\IEEEpeerreviewmaketitle

\section{Introduction}
\IEEEPARstart{D}{igital} museums and virtual exhibitions are becoming important infrastructures for cultural heritage preservation, access, and education. With 3D digitization and web-based visualization, artifacts can be explored as inspectable objects rather than static catalog entries. Meanwhile, vision--language models (VLMs) enable natural-language interaction with these visual assets. Together, these technologies point toward museum systems that can explain artifacts, not merely display them. Ancient Greek pottery is a representative testbed because its interpretation relies on vessel shape, painted scenes, production technique, chronology, and provenance.
\begin{figure}[t]
    \centering
    \includegraphics[width=\linewidth]{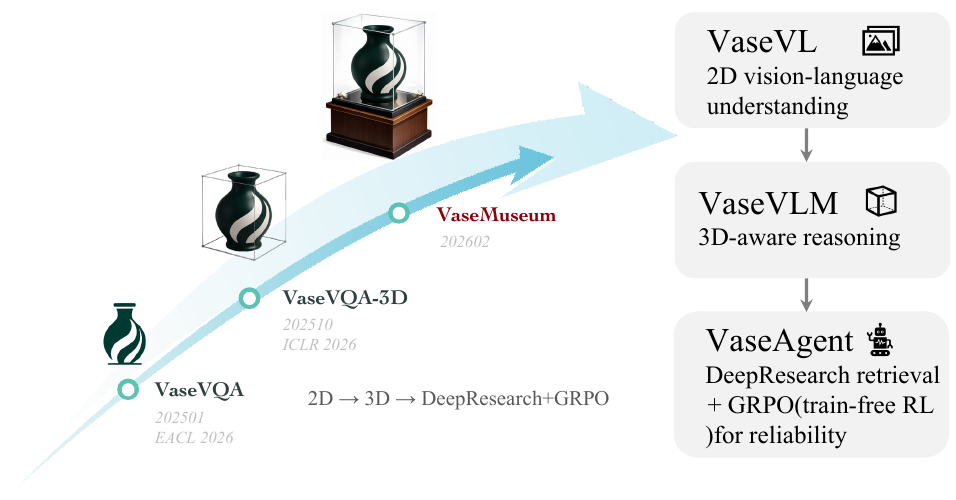}
    \caption{VaseVQA series development.}
    \label{fig:first-image-vase-series-development}
\end{figure}

\begin{figure*}[!ht]
    \centering
    \includegraphics[width=1.0\linewidth]{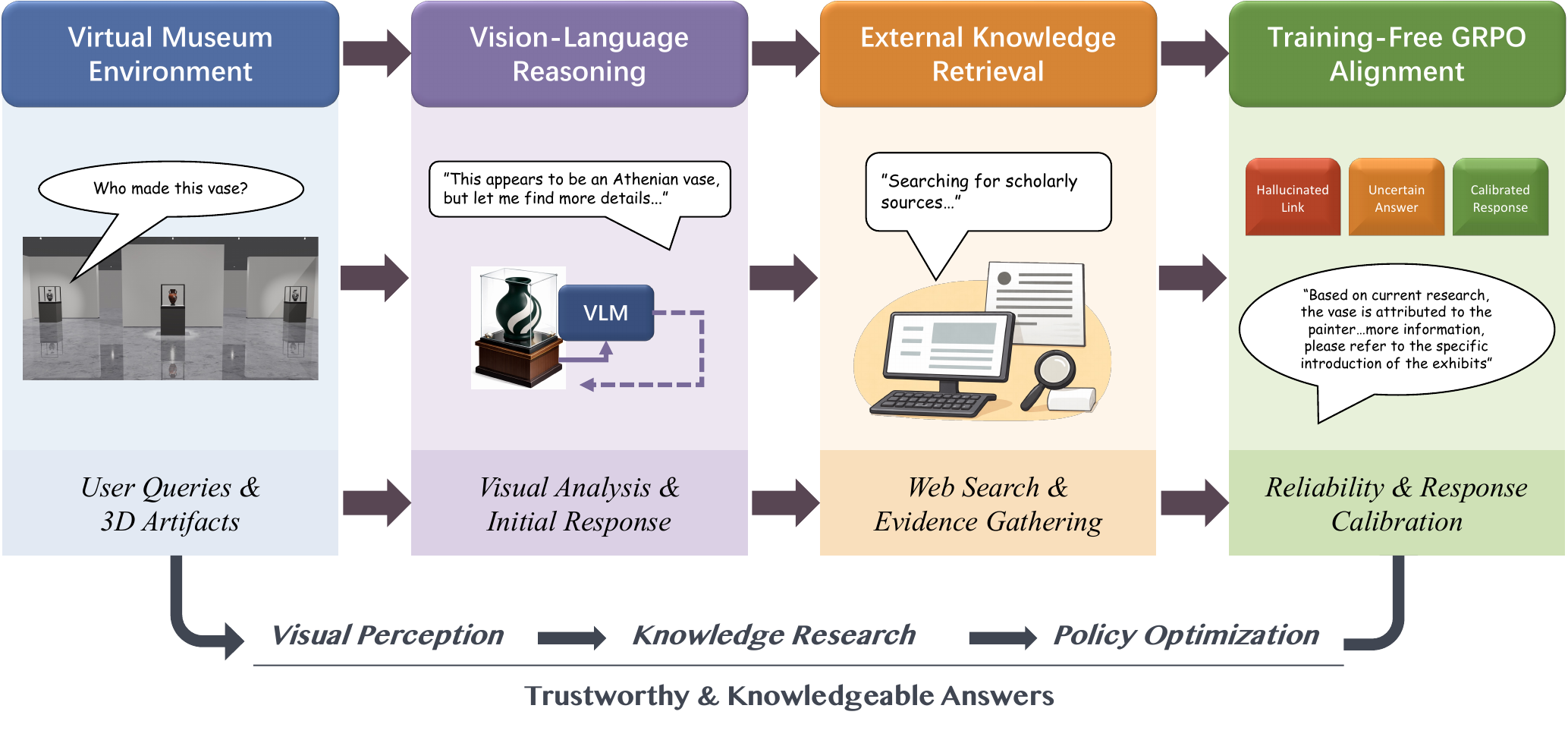}
    \caption{System architecture of VaseMuseum. The virtual museum provides artifact observations and user interactions, while VaseAgent integrates vision--language reasoning, external knowledge retrieval, source control, and response control to produce evidence-aware museum answers.}
    \label{fig:vasemuseum-architecture}
\end{figure*}
However, turning such systems into reliable VLM-based guides remains difficult. First, many museum questions cannot be answered from appearance alone; they require historical or archaeological evidence from external sources. Retrieval may then bring low-quality pages, incomplete records, or unverifiable citations into the context. Second, cultural-heritage evidence is often partial, disputed, or underspecified. As a result, generic VLMs may produce fluent but unsupported explanations when expert caution is needed. Our previous VaseVQA series, including VaseVQA and VaseVQA-3D~\cite{zhang2025vasevqa,ge2026vasevqa}, established domain-specific benchmarks for ancient Greek pottery and showed that domain-adaptive training improves structured vase understanding. Nevertheless, these benchmarks remain closer to closed-world evaluation than to real museum conversation, where users ask free-form questions, evidence quality varies, and reliable systems must decide when to answer, cite, or hedge.

To address these challenges, we introduce \textsc{VaseMuseum}, a lightweight multimodal agent framework for interactive digital museums of ancient Greek pottery. As shown in Fig.~\ref{fig:first-image-vase-series-development}, VaseMuseum extends our VaseVQA series from benchmark-style vase understanding to realistic museum interaction. It consists of a virtual museum interface and a reasoning agent, \textsc{VaseAgent}. The interface supports exploration of 2D images and 3D artifacts, while VaseAgent performs perception, 3D-aware reasoning, knowledge retrieval, and answer verification at inference time. Thus, VaseMuseum treats museum interaction as an evidence-seeking process rather than passive visualization.

Specifically, VaseAgent improves grounding through controlled evidence acquisition. When visual evidence is insufficient, it gathers information from authoritative web resources and museum knowledge sources through a DeepResearch-style tool loop. A source-control layer then suppresses unreliable hits and forms a compact, diverse evidence pool before generation. Meanwhile, response-level control improves uncertainty handling by comparing the answer with the collected evidence. Unsupported claims are discouraged, and insufficient or conflicting evidence leads to neutral, evidence-bounded responses. In addition, a training-free GRPO-style selection mechanism prefers candidates with more reliable references and better confidence behavior, without updating the VLM backbone~\cite{cai2025training,snell2024scaling,tan2025gtpo}.

We evaluate VaseMuseum in virtual museum interactions with visual-only, visual-plus-knowledge, and ambiguous questions. These tasks measure answer accuracy, link validity, hallucination, and neutrality under ambiguity. Compared with search-enabled VLM baselines, VaseMuseum improves link validity, lowers hallucination on knowledge-intensive queries, and yields more neutral responses under ambiguous evidence. Qualitative examples further show that the agent avoids fabricated or over-specific explanations by grounding claims in retrieved evidence. These results indicate that inference-time reliability control is a practical complement to stronger visual recognition for cultural-heritage deployment.

In summary, this paper studies trustworthy multimodal assistance for open-ended cultural-heritage interaction. Instead of relying only on larger VLMs or closed-set benchmark training, VaseMuseum controls both the evidence entering the context and the caution expressed in the final answer. The main contributions are as follows:

\begin{enumerate}
    \item We propose \textsc{VaseMuseum}, an intelligent digital museum framework that connects interactive 2D/3D artifact exploration with multimodal agent reasoning for ancient Greek pottery.
    \item We develop \textsc{VaseAgent}, an inference-time reasoning agent that combines DeepResearch-style knowledge acquisition with source-level control for more reliable evidence grounding.
    \item We introduce response-level reliability control and a training-free GRPO-style selection mechanism for uncertainty-aware answering, and validate their effectiveness against strong VLM baselines in realistic museum simulations.
\end{enumerate}

\section{Related Work}

\begin{figure*}[!ht]
    \centering
    \includegraphics[width=1.0\linewidth]{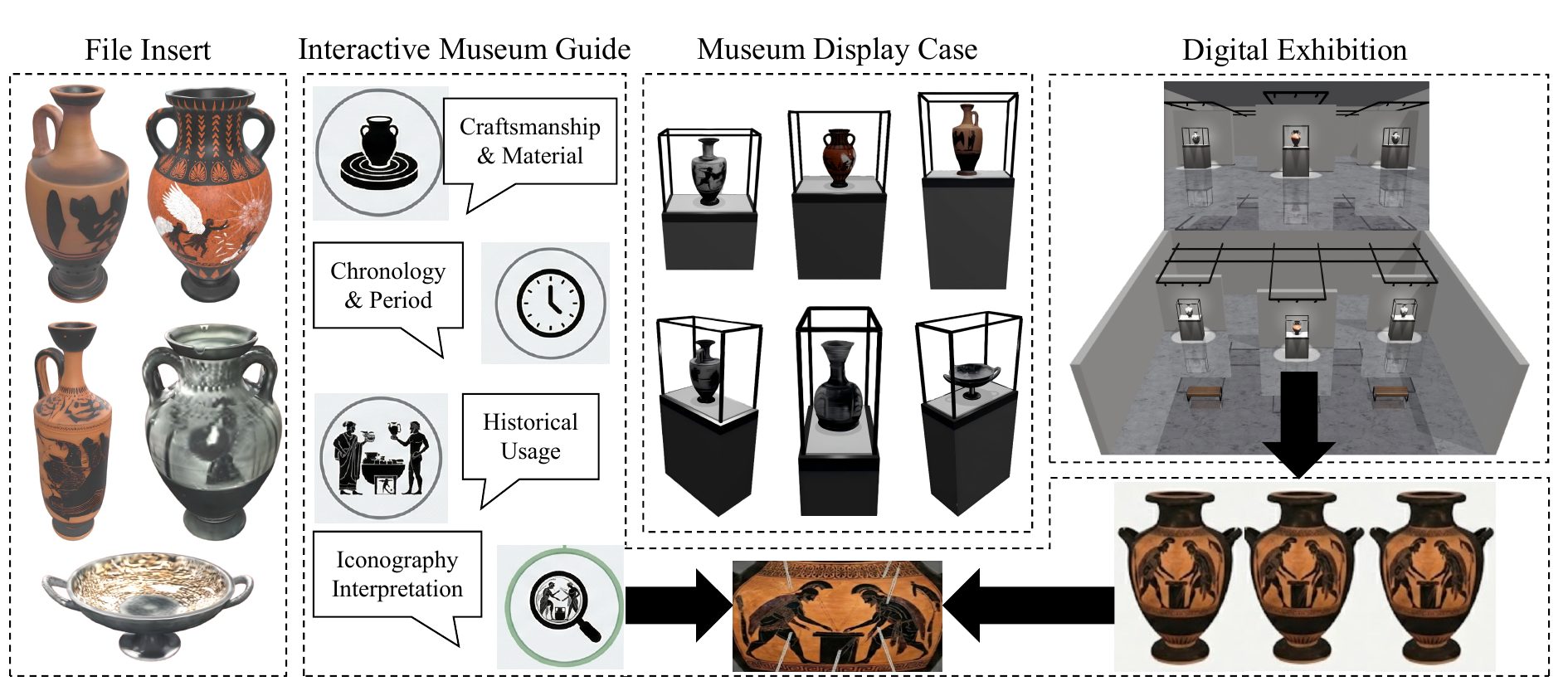}
    \caption{Virtual museum interaction workflow. After a user selects an exhibit and poses a question, VaseAgent analyzes the artifact observation, invokes external tools when visual evidence is insufficient, and applies reliability control before returning the final response.}
    \label{fig:vasemuseum_workflow}
\end{figure*}

\textbf{General Vision–Language Models and VQA.}
Modern large VLMs (e.g., GPT-4V, BLIP) excel on broad benchmarks like VQAv2 and image captioning\cite{zhang2025vasevqa}, but they rely on training data that rarely covers rare cultural objects. In practice, standard VQA datasets and models lack the deep domain context needed for art and artifacts: e.g., recent analysis shows that existing VQA benchmarks “often fail” to probe semantic understanding in art and cultural domains\cite{alfarano2025vqart,vitaloni2025comparative}. Similarly, prior work (e.g. VaseVQA-3D\cite{ge2026vasevqa}) finds that VLMs succeed on everyday images yet “struggle” on specialized 3D artifacts without targeted data or knowledge. VaseMuseum differs by embracing an interactive, open-domain interface over 3D artifacts, rather than only curating static datasets or retraining models for that domain.

\textbf{3D-Aware Vision–Language Methods.}
Recently, researchers have begun adding 3D awareness to VLMs. For example, Cap3D constructs a massive caption dataset by rendering 3D objects into multiple 2D views and captioning them with image-language models\cite{luo2023scalable,luo2024view}. Follow-up work like DiffuRank learns to select the most informative views (reducing hallucination) and refines 3D captions via large LMs\cite{huang20253d}. Other approaches embed spatial geometry directly: LLaVA-3D augments a 2D vision-language model with 3D positional embeddings, enabling a unified architecture that outputs 3D spatial understanding (like bounding boxes)\cite{zhu2025llava,huang20253drs,qi2024shapellm,yang2025prometheus}. While these methods advance 3D captioning and understanding, they generally rely on large-scale pretraining or finetuning on synthetic data. In contrast, VaseMuseum leverages off-the-shelf VLMs at inference time with planning and reliability controls, avoiding heavy retraining.

\textbf{Knowledge-Augmented Reasoning Agents.}
A parallel line of work explores LLM agents that augment vision with web and tool use. ReAct-style agents interleave chain-of-thought reasoning with explicit actions (e.g., API calls), allowing an LLM to query knowledge bases or browsers as it answers a question\cite{yao2022react,xi2025survey,li2025search,jin2025search}. In WebGPT, GPT-3 is fine-tuned to browse text information, collecting citations to improve factuality\cite{nakano2021webgpt}. More recently, the “deep research” paradigm formally combines planning, query generation, and retrieval: agents decompose a high-level query into subtasks, issue web searches, and synthesize evidence into answers\cite{zhang2025deep,huang2025deep,peng2026mta,su2026miroflow,chen2026jade}. State-of-the-art multimodal agents like Alibaba’s WebWatcher extend this to images, combining vision and text during web exploration for visual QA\cite{zhang2025deep,geng2025webwatcher,chen2025medbrowsecomp,ashraf2025agent}. Our method similarly uses a planning agent and external search to ground answers, but it is tailored to museum artifacts: crucially, VaseMuseum adds deterministic source and response control layers around each search round and the final answer, improving faithfulness without gradient updates on the VLM, distinguishing it from methods that simply scale data or tool use.

\textbf{Cultural Heritage and Digital Museum Systems.}
In cultural heritage, there is growing interest in digital replicas and AI-guided exhibits. “Digital museums” – virtual extensions of physical collections – now include interactive 3D reconstructions and VR exhibitions. Taking online access as the core carrier and interactive experience as an important support, digital museums break the temporal and spatial limitations of traditional museums\cite{parry2005digital,shiaw20043d}. Generative and AI technologies support these: for example, recent work highlights AI-driven restoration, 3D reconstruction, and multimodal storytelling in virtual exhibits\cite{xu2025review}. They not only provide the public with convenient and diverse immersive visiting paths, but also effectively expand the coverage of knowledge dissemination, improve the efficiency and quality of knowledge dissemination, and realize the extensive dissemination and in-depth popularization of cultural heritage knowledge\cite{srinivasan2009digital,zheng2024influences,marty2008museum}. Projects like the Hangzhou Museum’s “Everyone is a Curator” empower visitors to plan personalized virtual exhibitions from digital artifacts\cite{xu2025review,goyal2026scholarpeer}. However, most existing systems remain either passive displays or demand offline preparation (e.g., curated VR tours or chatbot guides). VaseMuseum synthesizes these ideas by placing an open-domain VQA agent inside a virtual gallery: visitors can freely explore 3D pottery and ask questions that the system answers (with web evidence) in real time, a setting not addressed by prior dataset-focused or statically programmed cultural heritage demos.

In summary, VaseMuseum uniquely combines three strands: it operates in the open world of a virtual museum (unlike closed benchmark VQA or captioning tasks), it handles complex 3D artifact imagery without extensive retraining (unlike recent 3D-aware VLMs that rely on large specialized datasets), and it integrates internet-scale knowledge retrieval with inference-time trust measures (unlike prior museum systems or LLM agents that either produce unchecked answers or depend on costly finetuning). This synthesis addresses the “capability gap” of general VLMs on heritage data by embedding interactive reasoning and reliability checking into the museum experience\cite{zhang2025vasevqa,xie2024osworld,ullrich2025openapps}.

\section{The Proposed Method}
\label{sec:method}

\subsection{Overview}
\label{sec:method-overview}

VaseMuseum is an inference-time multimodal agent framework for trustworthy interaction with ancient Greek pottery in a virtual museum. As shown in Fig.~\ref{fig:vasemuseum-architecture}, it consists of four modules: virtual museum interaction, vision--language reasoning, external knowledge acquisition, and inference-time reliability control. Given a user query $q$ and an exhibit observation $I$, the vision--language backbone grounds the query in pottery-specific visual evidence. When visual evidence is insufficient, a DeepResearch-style retrieval loop acquires external knowledge from web and museum sources. Source-level and response-level controls then filter retrieved evidence and calibrate generated claims. The framework remains modular and training-free; a GRPO-style group-relative selector can be enabled when additional inference budget is available.

\begin{figure}[ht]
    \centering
    \includegraphics[width=\linewidth]{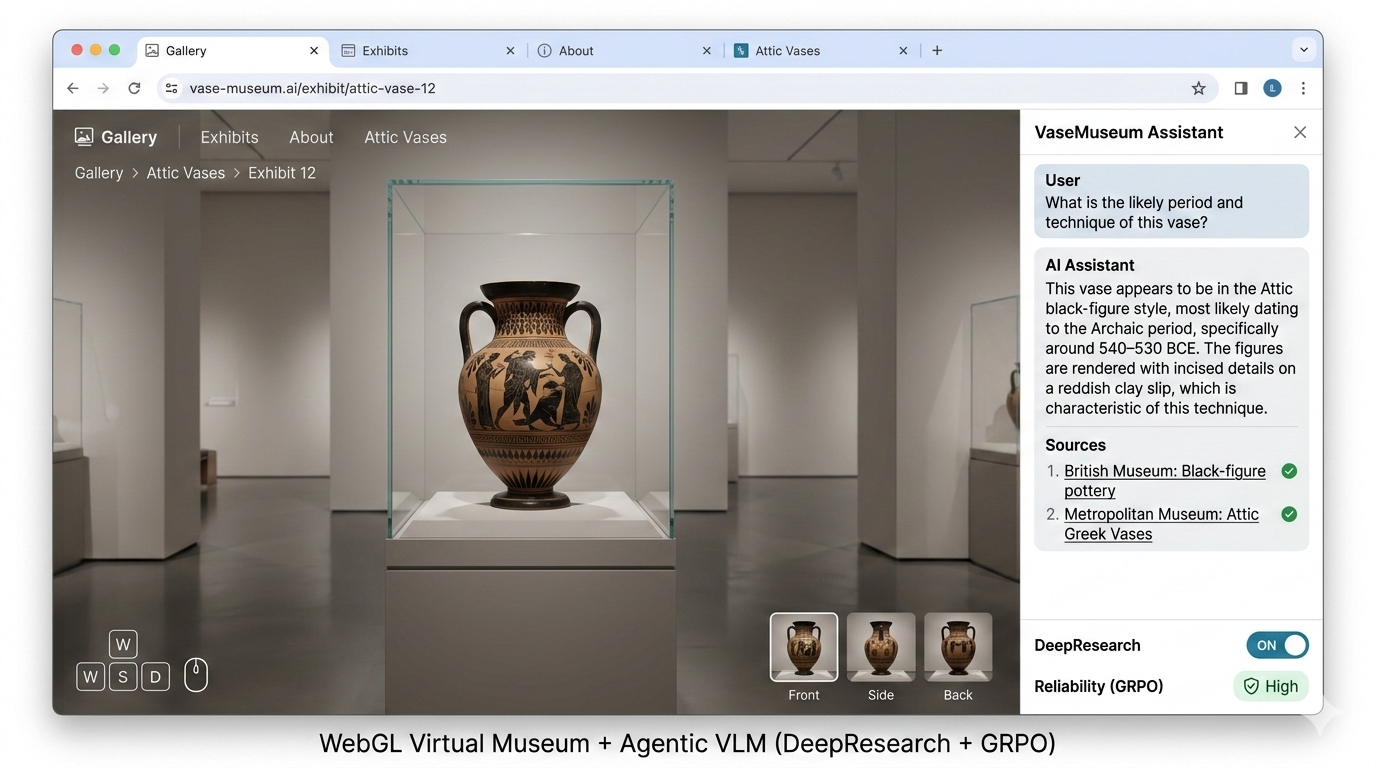}
    \caption{Virtual museum interface. Users can inspect ancient Greek pottery exhibits, select target artifacts, and ask natural-language questions; VaseAgent then returns explanations grounded in visual evidence and retrieved cultural-heritage knowledge.}
    \label{fig:vasemuseum-demo}
\end{figure}

\subsection{Virtual Museum Interaction and VaseAgent Workflow}
\label{sec:agent-workflow}

The virtual museum serves as the interaction environment for VaseAgent, with an example interface shown in Fig.~\ref{fig:vasemuseum-demo}. Each exhibit is represented by one or more visual observations, including high-resolution images or rendered views, and can be extended to 3D meshes when available. Users navigate the exhibition space, select an artifact, inspect visual details, and ask open-ended questions about form, technique, iconography, chronology, provenance, or cultural interpretation.

This environment is designed as an open-world setting rather than a closed VQA benchmark. As summarized in Fig.~\ref{fig:vasemuseum_workflow}, a museum visitor may ask questions whose answers depend on external scholarship, disputed attribution, or missing records. Accordingly, VaseAgent supports three response modes in this environment: answering from visual evidence, invoking external tools for knowledge acquisition, and producing a cautious response when evidence is insufficient.

\textbf{Multimodal reasoning and retrieval.}
VaseAgent serves as the reasoning core of VaseMuseum, coordinating visual understanding and external evidence acquisition. Given $(q,I)$, the vision--language backbone extracts pottery-specific visual cues, including vessel morphology, painting technique, depicted figures, scene composition, and iconographic motifs. These cues support visually answerable questions and provide search anchors for knowledge-dependent questions.

For questions that require information beyond visible appearance, VaseAgent switches from direct answering to tool-based evidence seeking. Specifically, the agent decomposes the user request into targeted search intents, issues web or museum-source queries, processes the returned snippets or pages, and fuses the results with the visual interpretation. Unlike single-shot retrieval-augmented generation, this retrieval process is iterative and conditioned on the agent's intermediate reasoning. As a result, the agent can refine the search when early results are incomplete, overly generic, or irrelevant to the selected artifact.

Retrieved evidence is not injected into the context without verification. Instead, each search payload is passed through source control before it is appended back to the dialogue context. After the tool loop stops, response control audits the merged evidence pool and determines whether the final answer should provide a direct response, hedge uncertain claims, or abstain from unsupported parts. Algorithm~\ref{alg:vaseagent-online} summarizes this single-trajectory inference process. Consequently, VaseAgent treats external knowledge as evidence to be checked rather than text to be copied.

\begin{figure}[t]
\centering
\rule{\linewidth}{0.4pt}
\footnotesize
\begin{algorithmic}[1]
\STATE \textbf{Input:} query $q$, observation $I$, VLM $\mathcal{M}$, rounds $R_{\max}$
\STATE \textbf{Output:} answer $a$
\STATE $\mathcal{D}\leftarrow(q,I)$; $\mathcal{E}\leftarrow\emptyset$; $\tilde{a}\leftarrow\emptyset$
\FOR{$t=1$ to $R_{\max}$}
    \STATE Sample $m_t\sim\mathcal{M}(\mathcal{D})$; append $m_t$ to $\mathcal{D}$
    \IF{$m_t$ contains no tool request}
        \STATE $\tilde{a}\leftarrow m_t$; \textbf{break}
    \ENDIF
    \FOR{each tool request in $m_t$}
        \STATE Execute tool and obtain payload $y$
        \STATE $\mathcal{E}_t\leftarrow\mathtt{SourceControl}(q,y)$
        \STATE $\mathcal{E}\leftarrow\mathtt{Merge}(\mathcal{E},\mathcal{E}_t)$
        \STATE Append $\mathcal{E}_t$ to $\mathcal{D}$
    \ENDFOR
\ENDFOR
\IF{$\tilde{a}=\emptyset$}
    \STATE Generate final draft $\tilde{a}\sim\mathcal{M}(\mathcal{D})$
\ENDIF
\STATE $g\leftarrow\mathtt{ResponseControl}(q,\tilde{a},\mathcal{E})$
\STATE $a\leftarrow\mathtt{Calibrate}(\tilde{a},g)$
\STATE \textbf{return} $a$
\end{algorithmic}
\rule{\linewidth}{0.4pt}
\vspace{0.4em}
\caption{\textbf{Pseudo-code of online VaseAgent inference.} The agent alternates between VLM reasoning and tool execution; each search payload is filtered by source control before entering the dialogue context, and the final draft is calibrated by response control.}
\label{alg:vaseagent-online}
\end{figure}

\subsection{Evidence Reliability Control}
\label{sec:evidence-control}

External search expands the agent's knowledge coverage but may introduce inaccessible URLs, duplicate snippets, low-quality sources, and unsupported claims. We therefore place two lightweight reliability controls around the retrieval loop. Source control filters and normalizes search results before they enter the dialogue context. Response control then audits the draft answer against the merged evidence pool before it is returned.

\textbf{Source control.}
Given a retrieved hit $h$, source control performs validity filtering, relevance--quality scoring, diversity-aware selection, and evidence normalization. The validity filter uses an accessibility proxy $a(h)$ from the URL and a text-sufficiency score $c(h)$ from the title and snippet. As defined in Eq.~\eqref{eq:source-validity}, a hit is retained only when
\begin{equation}
\begin{aligned}
    s_{\mathrm{val}}(h) &= 0.55\,a(h)+0.45\,c(h), \\
    h\in\mathcal{H}_{\mathrm{val}} &\Longleftrightarrow
    s_{\mathrm{val}}(h)\ge 0.40\ \text{and}\ a(h)\ge 0.50 .
\end{aligned}
\label{eq:source-validity}
\end{equation}
This step removes malformed links, unsafe schemes, and snippets that are too short for downstream verification.

For the remaining hits, we compute the pre-diversity score in Eq.~\eqref{eq:source-composite}, which combines query relevance, source quality, and accessibility. Let $\mathrm{rel}(h)$ be the token-level relevance between the query and hit text, $\rho_{\mathrm{site}}(h)$ be a domain prior, and $q_{\mathrm{src}}(h)=0.55\rho_{\mathrm{site}}(h)+0.45c(h)$ be the source-quality score:
\begin{equation}
\begin{aligned}
    \phi_{\mathrm{pre}}(h) ={}& 0.35\,\mathrm{rel}(h)
    +0.25\,q_{\mathrm{src}}(h) \\
    &+0.20\,\bar{a}(h),
    \qquad
    \bar{a}(h)=\frac{a(h)+c(h)}{2} .
\end{aligned}
\label{eq:source-composite}
\end{equation}
We discard hits with $\phi_{\mathrm{pre}}(h)<\tau_{\mathrm{keep}}$, where $\tau_{\mathrm{keep}}=0.22$.

To reduce redundancy, source control selects at most $N_{\mathrm{src}}=5$ hits by maximal marginal relevance. Given the selected set $S$, define $\mathrm{div}(h,S)=1-\max_{s\in S}\mathrm{Jac}(h,s)$, where $\mathrm{Jac}(h,s)$ is the Jaccard similarity between hit texts. The greedy objective in Eq.~\eqref{eq:mmr-objective} is
\begin{equation}
\begin{aligned}
    \mathrm{MMR}(h,S) ={}& \lambda_{\mathrm{MMR}}\,\mathrm{rel}(h)
    +(1-\lambda_{\mathrm{MMR}})\,\mathrm{div}(h,S) \\
    &+\beta_{\mathrm{dom}}(h,S)+0.04\,\rho_{\mathrm{site}}(h),
\end{aligned}
\label{eq:mmr-objective}
\end{equation}
where $\lambda_{\mathrm{MMR}}=0.72$ and $\beta_{\mathrm{dom}}(h,S)$ rewards a previously unseen domain. Each selected hit is serialized as a normalized evidence record containing its source identifier, URL, domain, modality, fused text, and diagnostic scores.

\textbf{Response control.}
Response control evaluates whether the draft answer $\tilde{a}$ is supported by the evidence pool $\mathcal{E}=\{e_1,\ldots,e_{|\mathcal{E}|}\}$. It first decomposes the user question and $\tilde{a}$ into coarse claim units using rule-based sentence and clause delimiters. For each claim $c_i$ and evidence item $e_j$, lexical overlap is computed; $e_j$ is marked as supporting $c_i$ when the overlap exceeds $\tau_{\mathrm{sup}}=0.08$. This yields a sparse claim--evidence matrix for coverage estimation.

Response control also records potential cross-source conflicts. For each evidence pair, it checks whether the two records have high lexical overlap but asymmetric negation or disagreement cues. Such pairs reduce confidence because cultural-heritage records often differ in dating, attribution, and interpretation.

\begin{figure*}[ht]
    \centering
    \includegraphics[width=\linewidth]{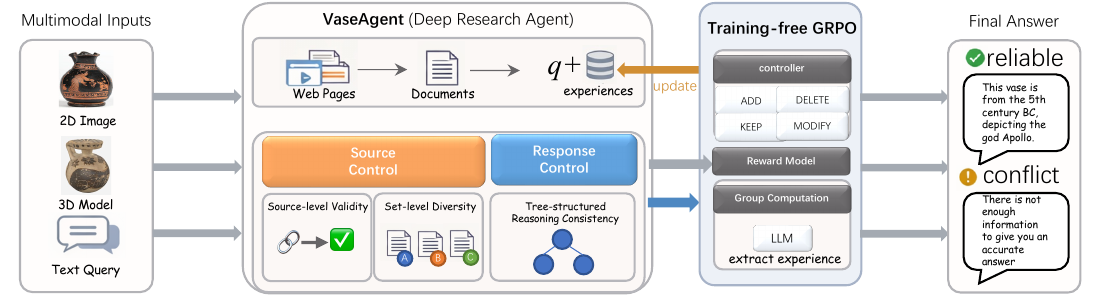}
    \caption{VaseAgent reliability framework. Source control removes invalid, low-quality, and redundant search results before context insertion, while response control checks generated claims against the merged evidence pool and triggers uncertainty-aware answering when support is insufficient or conflicting.}
    \label{fig:vaseagent-framework}
\end{figure*}

Figure~\ref{fig:vaseagent-framework} illustrates the reliability-control pipeline. Finally, the response confidence score in Eq.~\eqref{eq:response-confidence} combines claim coverage, consistency, multi-source support, and conflict penalties:
\begin{equation}
\begin{aligned}
    \psi ={}& 0.40\,c_{\mathrm{cov}}
    +0.30\,\kappa_{\mathrm{cons}}
    +0.20\,\omega_{\mathrm{multi}}
    +0.10\,m_{\mathcal{E}} \\
    &-\min\bigl(0.35,\,0.12\,n_{\mathrm{conf}}\bigr),
    \qquad
    \psi\leftarrow\mathrm{clip}(\psi,0,1).
\end{aligned}
\label{eq:response-confidence}
\end{equation}
Here, $c_{\mathrm{cov}}$ is the fraction of claims with supporting evidence, $\kappa_{\mathrm{cons}}$ penalizes unsupported claims, $\omega_{\mathrm{multi}}$ is the fraction of supported claims backed by multiple sources, $m_{\mathcal{E}}$ indicates whether the evidence pool is non-empty, and $n_{\mathrm{conf}}$ is the number of detected conflict pairs. If $\psi<\tau_{\mathrm{conf}}$ with $\tau_{\mathrm{conf}}=0.52$, VaseAgent enters uncertain mode and prepends an evidence-bounded preamble that identifies unsupported or conflicting aspects; otherwise, it returns the answer in normal mode and logs the audit record for evaluation.

\subsection{Training-Free GRPO-Style Response Selection}
\label{sec:training-free-grpo}

The default online path in Algorithm~\ref{alg:vaseagent-online} uses a single controlled trajectory for efficiency. When additional inference budget is available, VaseMuseum further applies a training-free Group Relative Policy Optimization (GRPO)-style selector without updating the VLM backbone. This selector exploits variation among stochastic tool-augmented rollouts, which may differ in URL validity, evidence coverage, and uncertainty calibration.

Given $(q,I)$, the selector samples $K$ independent controlled trajectories $\{\tau_k\}_{k=1}^{K}$ and obtains corresponding answers $\{a_k\}_{k=1}^{K}$. Each candidate is scored by lightweight verifiers for link validity, claim--evidence support, neutrality under insufficient evidence, and conflict awareness. The final answer is selected by the group-relative objective in Eq.~\eqref{eq:group-selection}
\begin{equation}
\begin{aligned}
    k^* &= \arg\max_{1\le k\le K} R(a_k,\tau_k,q,I), \\
    a^* &= a_{k^*} .
\end{aligned}
\label{eq:group-selection}
\end{equation}
where $R(\cdot)$ denotes the aggregated reliability score. Since $R(\cdot)$ only reranks frozen-model outputs, the procedure remains training-free and can be disabled under strict latency constraints.

\begin{figure}[t]
\centering
\rule{\linewidth}{0.4pt}
\footnotesize
\begin{algorithmic}[1]
\STATE \textbf{Input:} $q$, $I$, VLM $\mathcal{M}$, group size $K$
\STATE \textbf{Output:} selected answer $a^*$
\FOR{$k=1$ to $K$}
    \STATE Run Algorithm~\ref{alg:vaseagent-online} with seed $k$
    \STATE Obtain $(\tau_k,\mathcal{E}_k,a_k)$
    \STATE Compute $R(a_k,\tau_k,q,I)$
\ENDFOR
\STATE $k^*\leftarrow\arg\max_k R(a_k,\tau_k,q,I)$
\STATE \textbf{return} $a_{k^*}$
\end{algorithmic}
\rule{\linewidth}{0.4pt}
\vspace{0.4em}
\caption{\textbf{Pseudo-code of training-free GRPO-style response selection.} When additional inference budget is available, multiple controlled trajectories are sampled with different seeds and reranked by reliability verifiers for link validity, evidence support, neutrality, and conflict awareness, without updating the VLM.}
\label{alg:grpo}
\end{figure}

Algorithm~\ref{alg:grpo} gives the corresponding pseudo-code. Beyond online selection, the same grouped rollouts can be used offline to collect high-reliability prompt and evidence patterns. These textual experiences may guide later deployment but are not required by the default single-trajectory agent.

\subsection{Evaluation Data Construction}
\label{sec:data-construction}

We construct the evaluation set from the curated image collection of VaseVQA-3D, which contains more than 3,000 ancient Greek vase images with structured annotations such as fabric, technique, shape, date, attribution, and decoration. To support knowledge-intensive queries, we identify 518 vases with valid links to the \textsc{LIMC} database (Lexicon Iconographicum Mythologiae Classicae)~\cite{limc2014weblimc}.

\textbf{Cross-source enrichment.}
Using LIMC identifiers as anchors, we align vase records with related scholarly resources, including museum catalogues and classical iconography databases when available. This enrichment adds iconographic descriptions, mythological figure annotations, archaeological context, alternative attributions, and bibliographic references. As a result, the evaluation set supports questions that require both visual observation and external cultural-heritage evidence.

\textbf{Task construction.}
We instantiate three task categories. Visual-only questions are answerable from the exhibit image, such as vessel shape or visible scene elements. Visual-plus-knowledge questions require combining visual cues with external information, such as dating, function, mythological interpretation, or cultural context. Ambiguous questions are underspecified or involve disputed records, and therefore evaluate whether a system can avoid overconfident claims when multiple answers are plausible.

\textbf{Generation and checking.}
For each vase, we combine rendered imagery, captions, iconographic notes, and enriched metadata to create LLM-assisted question templates. Annotators spot-check the generated instances for source fidelity, answerability, and task-type clarity. We balance the three evaluation splits at 100 instances each, and separately collect 100 paraphrases of held-out queries to refine prompts without leaking exact test items into the evaluation set.

\textbf{Design rationale.}
VaseAgent prioritizes inference-time reliability over additional end-to-end training. This design is suitable for cultural-heritage deployment because museum collections vary across institutions, authoritative sources evolve, and many artifacts have incomplete or contested records. The modular structure allows each component---perception, retrieval, source control, and response control---to be updated independently for new collections without retraining the full system.

The separation of perception, retrieval, and reliability control also improves inspectability. When a failure occurs, the audit log can indicate whether it originates from visual recognition, weak retrieval, source filtering, or answer calibration. This inspectability is important for museum scenarios, where citation validity and uncertainty awareness are as critical as answer accuracy.

\section{Experiments}
\label{sec:experiments}

\subsection{Experimental Setup}
\label{sec:exp-setup}

We evaluate \textsc{VaseAgent} in virtual museum interactions that require visual recognition, external knowledge grounding, and uncertainty-aware answering. Each evaluation instance contains an exhibit observation and a natural-language query. As summarized in Table~\ref{tab:exp-protocol}, all methods receive the same visual input and query; search-enabled methods may invoke external tools, while GRPO-style selection samples $K=4$ controlled trajectories unless otherwise specified.

\textbf{Metric computation.}
Answer accuracy is computed against reference annotations or acceptable answer sets for each query. For ambiguous questions, an answer is considered appropriate when it either matches a plausible annotated interpretation or explicitly states that the evidence is insufficient for a unique conclusion. Hallucination rate measures the fraction of responses that contain factual claims unsupported by the exhibit observation, structured metadata, or retrieved evidence. Groundedness measures whether the main answer claims are supported by visual evidence or valid external sources. Link validity counts cited URLs that are accessible and relevant to the generated claim. Neutrality is rated on a 0--5 scale, where higher scores indicate more cautious and evidence-bounded wording under ambiguity.

\begin{table}[t]
\centering
\caption{Evaluation protocol. Accuracy (A), groundedness (G), link validity, and neutrality are better when higher; hallucination (H) is better when lower.}
\label{tab:exp-protocol}
\scriptsize
\setlength{\tabcolsep}{3.2pt}
\begin{tabular}{p{0.23\linewidth}p{0.68\linewidth}}
\toprule
Item & Description \\
\midrule
Tasks & V-Only: visually answerable artifact questions; V+K: questions requiring external historical or archaeological knowledge; Amb.: underspecified or disputed questions with multiple plausible interpretations. \\
Metrics & A: answer accuracy; H: hallucination rate; G: groundedness score; link validity: fraction of valid external citations; neutrality: human-rated uncertainty handling on a 0--5 scale. \\
Scoring & Accuracy is checked against reference answers; hallucination and groundedness are judged against visual evidence, metadata, and retrieved sources; link validity requires accessible and relevant citations; neutrality rewards cautious wording when evidence is ambiguous. \\
Baselines & Qwen3-VL-8B (Direct): zero-shot answering without search; Qwen3-VL-8B + Search: search-enabled answering without reliability control; VaseAgent (w/o GRPO): source/response control without group-relative selection; VaseAgent (Full): complete system with GRPO-style selection. \\
\bottomrule
\end{tabular}
\end{table}

\subsection{Main Results}
\label{sec:main-results}

Table~\ref{tab:accuracy} reports performance across visual-only, visual-plus-knowledge, and ambiguous queries. Direct VLM answering remains competitive on visual-only and ambiguous recognition metrics, where short visual answers are often sufficient. However, search without reliability control increases hallucination and reduces groundedness.

In contrast, \textsc{VaseAgent} (Full) is strongest on knowledge-intensive reliability: it obtains the lowest hallucination rate on V+K queries and the highest groundedness and neutrality scores in the same setting. These results indicate that controlled evidence acquisition is most beneficial when artifact interpretation requires external knowledge rather than visual recognition alone.

\begin{table}[t]
\centering
\caption{Performance across query types. A, H, G, and N denote answer accuracy, hallucination rate, groundedness, and neutrality. Higher is better for A/G/N, lower is better for H, and best/second-best results are marked in bold/underline.}
\label{tab:accuracy}
\scriptsize
\setlength{\tabcolsep}{2.2pt}
\resizebox{\linewidth}{!}{
\begin{tabular}{lcccccccccccc}
\toprule
& \multicolumn{4}{c}{V-Only}
& \multicolumn{4}{c}{V+K}
& \multicolumn{4}{c}{Amb.} \\
\cmidrule(lr){2-5}
\cmidrule(lr){6-9}
\cmidrule(lr){10-13}
Method
& A $\uparrow$ & H $\downarrow$ & G $\uparrow$ & N $\uparrow$
& A $\uparrow$ & H $\downarrow$ & G $\uparrow$ & N $\uparrow$
& A $\uparrow$ & H $\downarrow$ & G $\uparrow$ & N $\uparrow$ \\
\midrule
Qwen3-VL-8B (Direct)
& \textbf{.44} & \textbf{.54} & \textbf{.45} & \textbf{2.83}
& \textbf{.69} & \underline{.29} & \underline{.70} & \underline{3.46}
& \textbf{.65} & \textbf{.33} & \textbf{.67} & \textbf{3.33} \\
Qwen3-VL-8B + Search
& .20 & .79 & .20 & 1.86
& \underline{.59} & .44 & .55 & 2.91
& .48 & .53 & .46 & 2.64 \\
VaseAgent (w/o GRPO)
& .20 & .76 & .24 & 2.18
& .53 & .38 & .60 & 3.04
& \underline{.55} & .46 & .54 & 2.84 \\
\textbf{VaseAgent (Full)}
& \underline{.24} & \underline{.66} & \underline{.34} & \underline{2.47}
& .54 & \textbf{.22} & \textbf{.73} & \textbf{3.75}
& \underline{.55} & \underline{.44} & \underline{.56} & \underline{3.02} \\
\bottomrule
\end{tabular}}
\end{table}

\begin{figure}[t]
    \centering
    \includegraphics[width=0.95\linewidth]{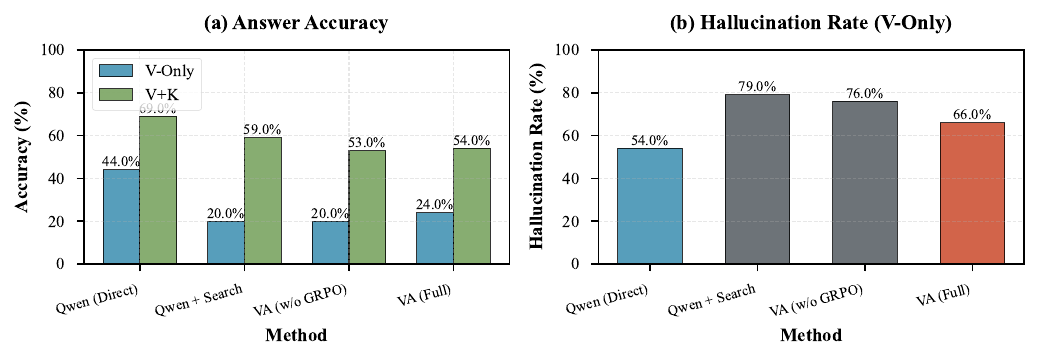}
    \caption{Main comparison across baselines. The figure contrasts answer accuracy and hallucination behavior across methods, highlighting that reliability control is most useful when external evidence is required.}
    \label{fig:main_comparison}
\end{figure}

Figure~\ref{fig:main_comparison} visualizes the same trend from the main comparison. Rather than uniformly improving all recognition-style metrics, VaseAgent primarily improves reliability-sensitive metrics under knowledge-intensive interaction. This distinction is important for digital museums, where unsupported but fluent explanations are more harmful than conservative answers.

\begin{figure}[t]
    \centering
    \includegraphics[width=0.95\linewidth]{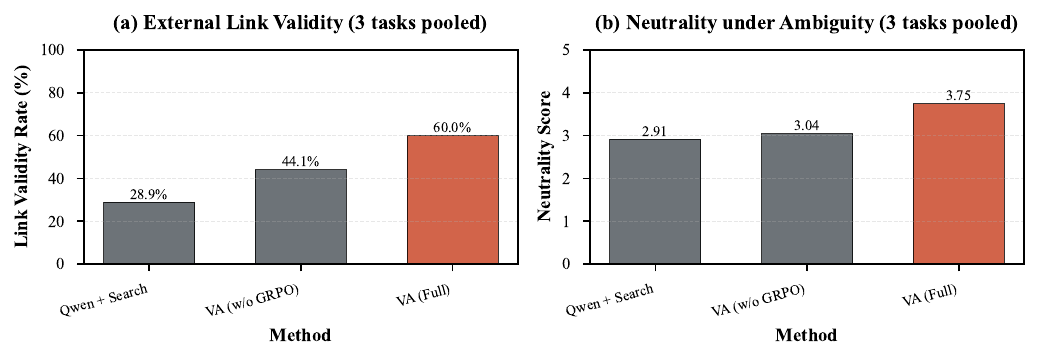}
    \caption{Reliability metrics. VaseAgent improves external link validity and neutrality under ambiguous queries by filtering retrieved sources and calibrating generated claims against the evidence pool.}
    \label{fig:reliability}
\end{figure}

\begin{table}[t!]
\centering
\caption{External link validity and neutrality under ambiguity. Link validity measures citation reliability, and neutrality measures cautious answering under ambiguous evidence.}
\label{tab:reliability}
\setlength{\tabcolsep}{4pt}
\begin{tabular}{lcc}
\toprule
Method & Link Validity $\uparrow$ & Neutrality $\uparrow$ \\
\midrule
Qwen3-VL-8B + Search & 28.89 & 2.91 \\
VaseAgent (w/o GRPO) & \underline{44.07} & 3.04 \\
\textbf{VaseAgent (Full)} & \textbf{60.00} & \textbf{3.75} \\
\bottomrule
\end{tabular}
\end{table}

\subsection{Reliability Analysis}
\label{sec:reliability-analysis}

Table~\ref{tab:reliability} isolates two deployment-oriented metrics: external link validity and neutrality under ambiguity. Search alone provides access to more information but produces many unreliable citations. Adding source and response control improves link validity from 28.89 to 44.07, and the full GRPO-style selector further increases it to 60.00. Neutrality follows the same pattern, indicating that reliability control discourages overconfident claims when evidence is incomplete or conflicting.

Figure~\ref{fig:reliability} further summarizes citation validity and neutrality. The improvement from uncontrolled search to VaseAgent indicates that evidence filtering and answer auditing are necessary complements to tool use in museum-oriented VLM systems.

\begin{figure*}[t]
    \centering
    \begin{subfigure}{0.32\linewidth}
        \centering
        \includegraphics[width=\linewidth]{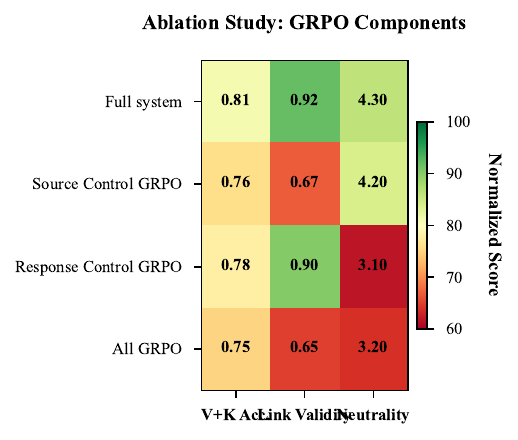}
        \caption{Ablation of reliability-control components. Source-side signals primarily affect citation validity, while response-side signals affect uncertainty-aware answering.}
        \label{fig:ablation}
    \end{subfigure}
    \hfill
    \begin{subfigure}{0.32\linewidth}
        \centering
        \includegraphics[width=\linewidth]{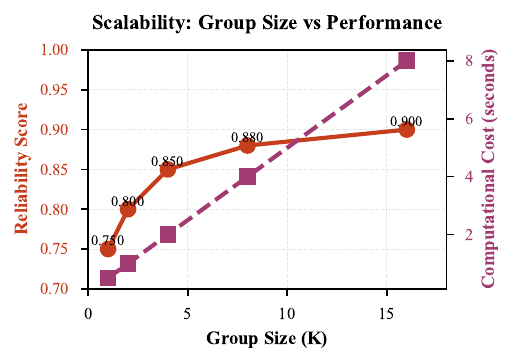}
        \caption{Scaling behavior of the GRPO-style selector. Reliability improves with group size but saturates beyond $K=4$, whereas inference cost grows approximately linearly.}
        \label{fig:scalability}
    \end{subfigure}
    \hfill
    \begin{subfigure}{0.32\linewidth}
        \centering
        \includegraphics[width=\linewidth]{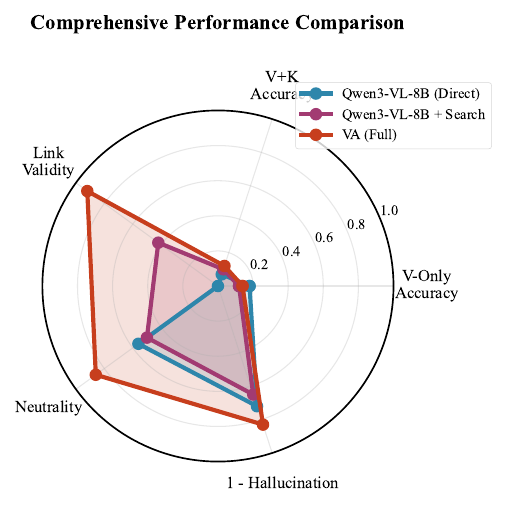}
        \caption{Multi-dimensional comparison across accuracy, groundedness, citation validity, neutrality, and hallucination reduction. VaseAgent provides a more balanced reliability profile for museum interaction.}
        \label{fig:radar}
    \end{subfigure}
    \caption{Ablation, scalability, and multi-dimensional evaluation of VaseAgent.}
    \label{fig:overall_analysis}
\end{figure*}

\subsection{Ablation and Scaling}
\label{sec:ablation-scaling}

Table~\ref{tab:ablation} studies how reliability-control components affect V+K accuracy, link validity, and neutrality. The full system achieves the best link validity and neutrality, while the variant denoted as All GRPO obtains higher V+K accuracy but lower reliability. This trade-off suggests that optimizing only for answer correctness can encourage more assertive responses, whereas source and response controls are needed to preserve citation quality and calibrated uncertainty.

\begin{table}[t]
\centering
\caption{Ablation of reliability-control components. The full system gives the strongest reliability profile, while accuracy-oriented selection can reduce citation validity and neutrality.}
\label{tab:ablation}
\resizebox{\linewidth}{!}{%
\begin{tabular}{lccc}
\toprule
Configuration & V+K Acc. $\uparrow$ & Link Validity $\uparrow$ & Neutrality $\uparrow$ \\
\midrule
Full system & \underline{54.00} & \textbf{60.00} & \textbf{3.75} \\
\quad w/o source-control reward & 53.00 & \underline{44.07} & \underline{3.04} \\
\quad w/o response-control reward & 53.61 & 33.33 & 2.77 \\
\quad accuracy-oriented GRPO & \textbf{59.00} & 28.89 & 2.91 \\
\bottomrule
\end{tabular}%
}
\end{table}

Figure~\ref{fig:ablation} presents the same ablation visually. Removing source-related reliability signals mainly harms link validity, whereas removing response-related signals reduces neutrality. Thus, the two controls address complementary failure modes: weak evidence entering the context and overconfident claims leaving the system.

Figure~\ref{fig:scalability} analyzes the effect of group size $K$ in the GRPO-style selector. Increasing $K$ improves reliability because the selector has more candidate trajectories to compare, but the gain saturates beyond $K=4$ while computational cost continues to grow approximately linearly. We therefore use $K=4$ as the default trade-off between reliability and efficiency.

\subsection{Task Breakdown and Qualitative Discussion}
\label{sec:qualitative-discussion}

\begin{figure}[t]
    \centering
    \includegraphics[width=0.95\linewidth]{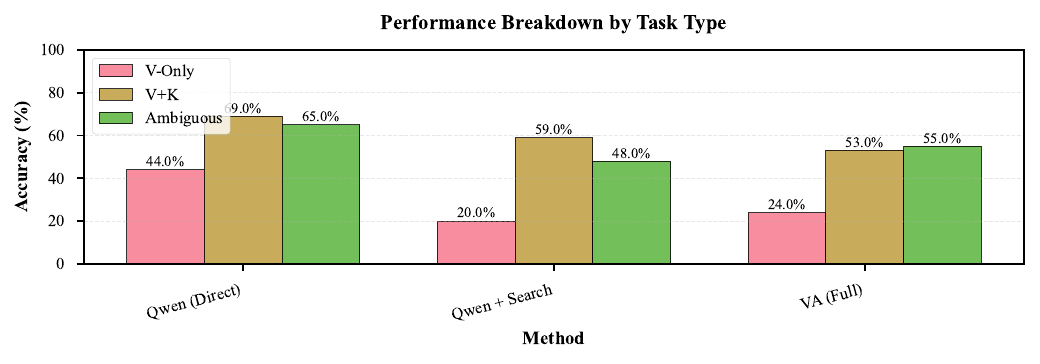}
    \caption{Task-specific performance breakdown across V-Only, V+K, and Ambiguous queries. VaseAgent is most effective in knowledge-intensive settings where external evidence must be retrieved and verified.}
    \label{fig:task_breakdown}
\end{figure}

Figure~\ref{fig:task_breakdown} breaks down performance by task type. The full system is most beneficial for V+K queries, where external evidence and citation control directly affect answer quality. On visual-only and ambiguous accuracy metrics, direct VLM answering remains strong, but it lacks the citation and uncertainty controls required for trustworthy museum deployment.

Qualitatively, uncontrolled search-enabled baselines often produce authoritative-sounding explanations with weak or invalid references. VaseAgent instead checks retrieved evidence before generation and calibrates the final answer when support is insufficient. This behavior is aligned with curatorial practice: a museum assistant should distinguish between visually observable attributes, externally supported facts, and uncertain interpretations.

Figure~\ref{fig:radar} summarizes the multi-dimensional comparison across accuracy, groundedness, link validity, neutrality, and hallucination reduction. The radar view emphasizes that VaseAgent is designed for a balanced reliability profile rather than a single closed-set accuracy score.

\section{Limitations and Future Work}
\label{sec:limitations}

VaseMuseum still has several limitations. First, its evidence quality depends on the availability, stability, and coverage of external web and museum sources, which can be incomplete or inconsistent for specialized archaeological records. Second, our current implementation focuses on ancient Greek pottery, and extending the framework to other artifact categories, languages, and curatorial conventions will require domain-specific validation. Third, reliability-oriented metrics such as link validity and neutrality involve operational definitions and human judgment; future work should therefore incorporate curated institutional knowledge bases, broader heritage collections, and larger-scale expert evaluation with clearer agreement analysis.

\section{Conclusion}
\label{sec:conclusion}

This paper presented \textsc{VaseMuseum}, an inference-time multimodal agent framework for trustworthy interaction with ancient Greek pottery in a virtual museum. By combining visual reasoning, DeepResearch-style external knowledge acquisition, source-level evidence filtering, response-level answer calibration, and training-free GRPO-style response selection, VaseMuseum targets the reliability challenges that arise in open-ended cultural-heritage dialogue. Experiments show that the framework is particularly beneficial for knowledge-intensive queries, improving citation validity, reducing hallucination, and encouraging more neutral answers under ambiguous evidence without updating the VLM backbone. These results suggest that inference-time reliability control is a practical direction for building museum assistants that are visually grounded, evidence-aware, and appropriately cautious.

\noindent \textbf{Acknowledgments.}
The authors used GPT5.5 (OpenAI) to polish the language of this paper. This work was supported by the Fundamental Research Funds for the Central Universities, Peking University.

\small
\bibliographystyle{IEEEtran}
\bibliography{reference}

\begin{IEEEbiography}[{\includegraphics[width=1in]{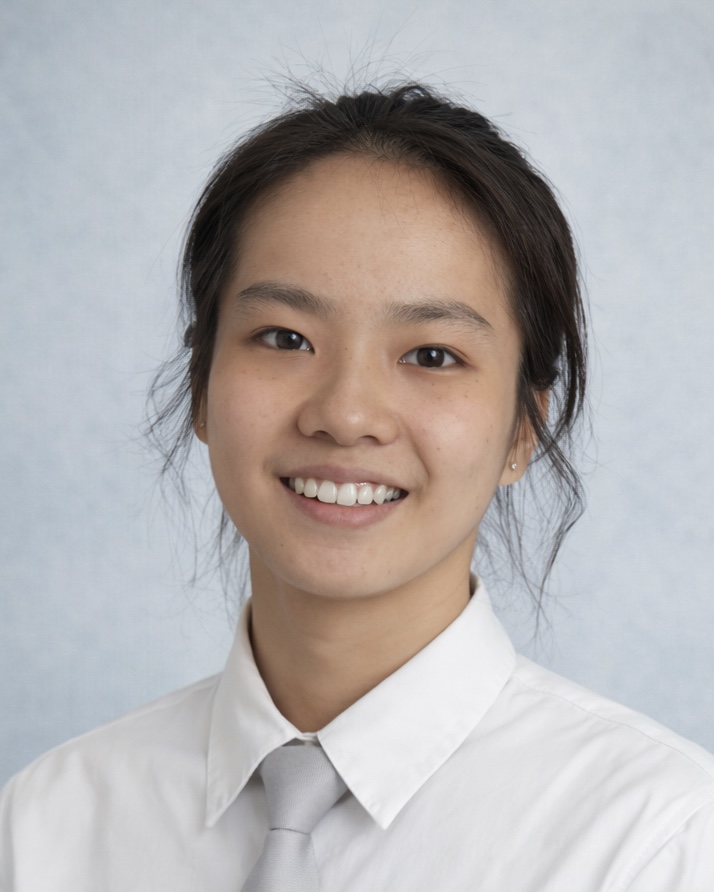}}]{Jiazi Wang}
is a second-year M.S. student at Beijing Jiaotong University, advised by Prof. Yufeng Chen. She received her B.S. degree from Beijing Jiaotong University. Her research interests include multilingual and multimodal large language models, cross-lingual knowledge modeling, and vision-language understanding, with applications to cultural heritage data analysis.
\end{IEEEbiography}

\begin{IEEEbiography}[{\includegraphics[width=1in]{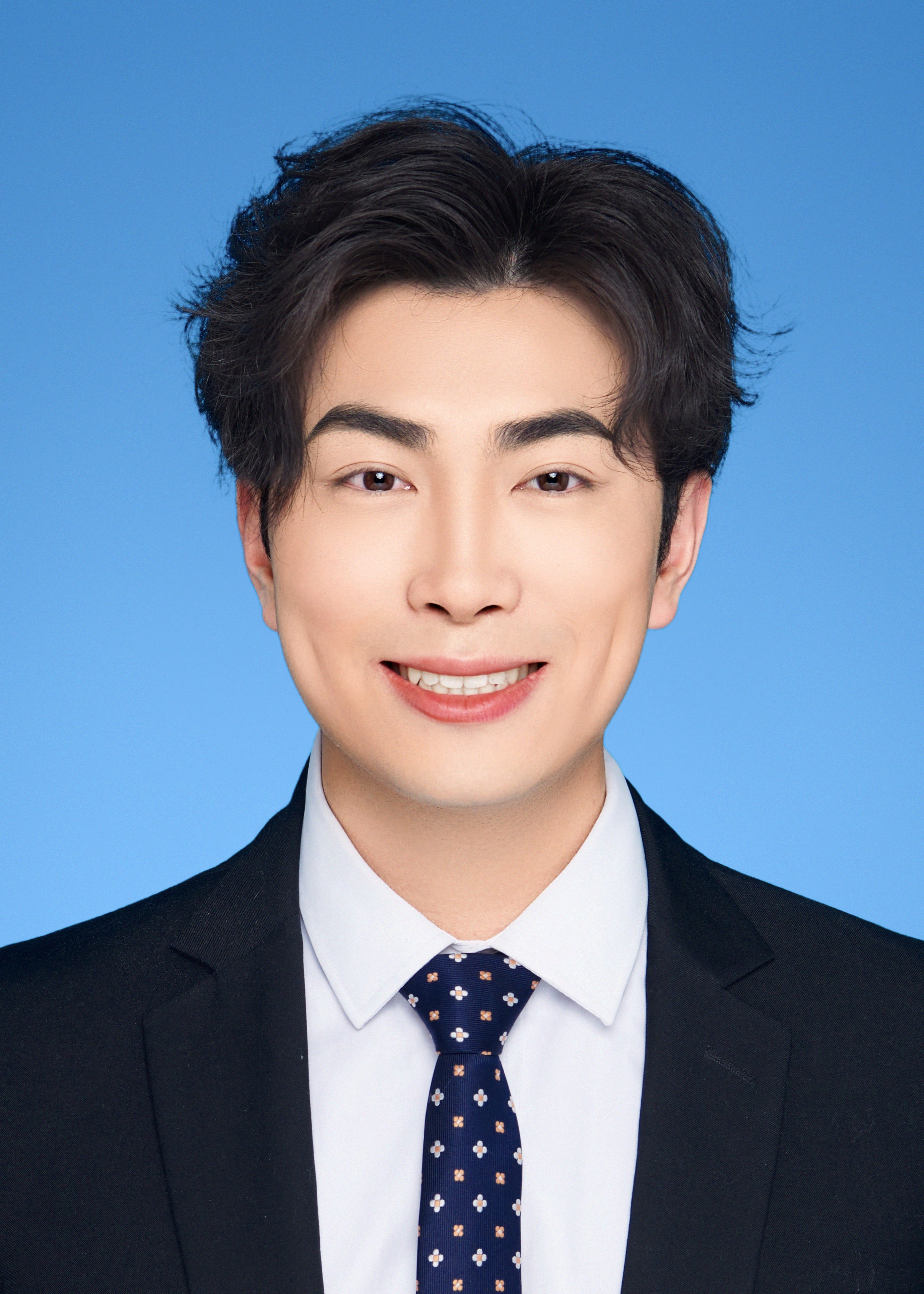}}]{Nonghai Zhang}
 is a second-year M.S. student of software engineering at Peking University. Received his B.S. in software engineering from Yun Nan University, Yunnan, China. His research interests include computer
vision, embodied AI, and LLM.
\end{IEEEbiography}

\begin{IEEEbiography}[{\includegraphics[width=1in]{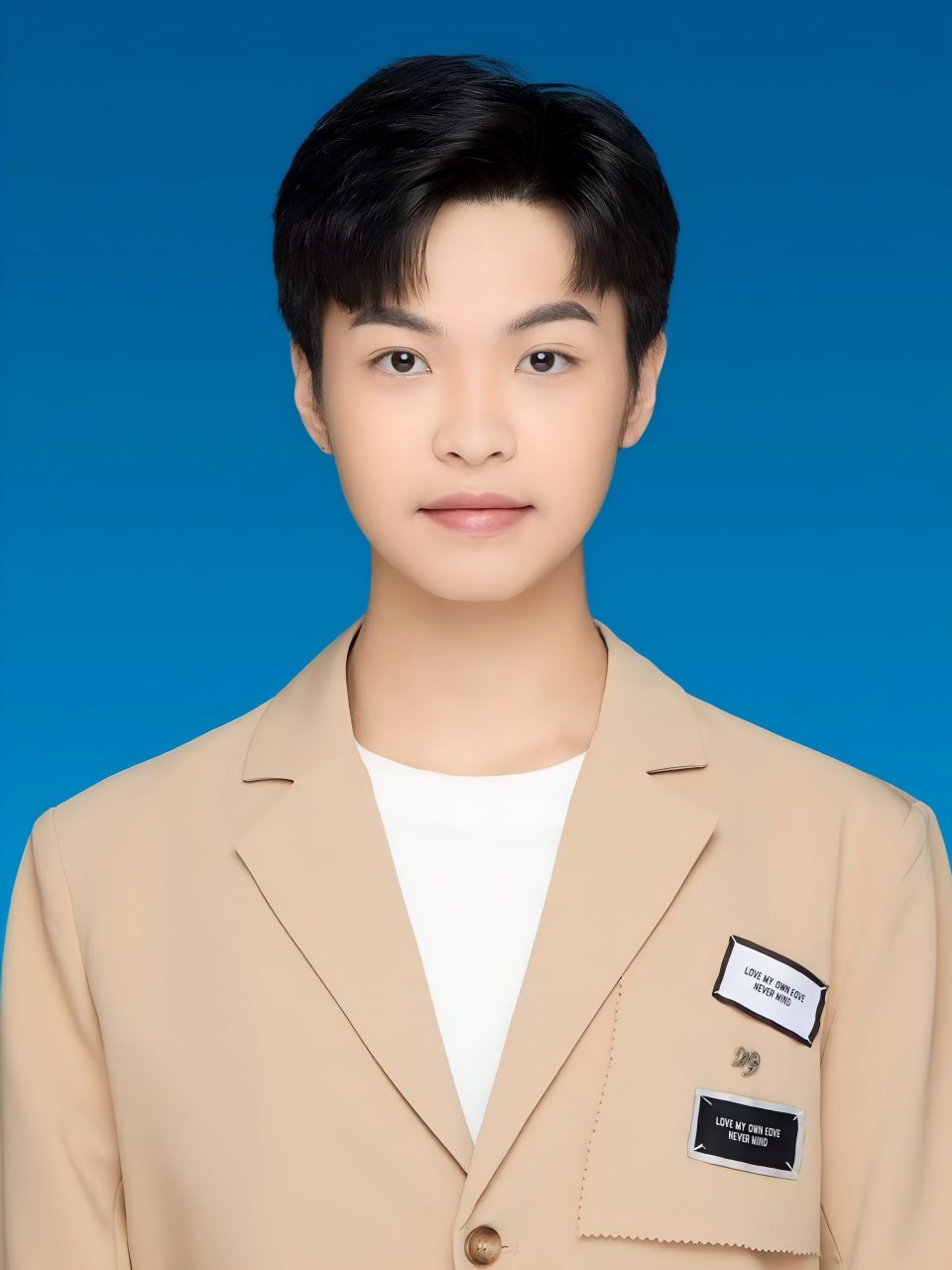}}]{Qiushi Xie}
is a third-year undergraduate student at Huazhong University of Science and Technology, Wuhan, China. His research interests lie in vision-language models, agentic systems, deep learning, and edge AI deployment, with applications to interactive digital twins.
\end{IEEEbiography}

\begin{IEEEbiography}[{\includegraphics[width=1in]{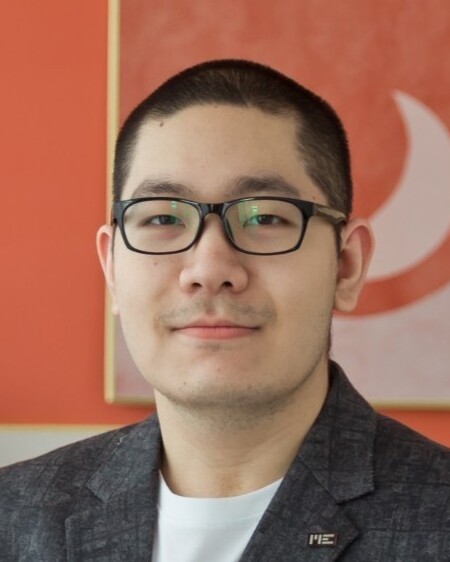}}]{Zeyu Zhang}
is an incoming PhD student at UC Berkeley BAIR, advised by Prof. Pieter Abbeel, Prof. Alexei A. Efros, and Prof. Angjoo Kanazawa. He received his bachelor’s degree from the Australian National University, where he was advised by Prof. Richard Hartley and Prof. Ian Reid. His research interests lie in geometric generative modeling and its applications to world models, multimodal foundation models, embodied AI, and AI for health. 
\end{IEEEbiography}

\begin{IEEEbiography}[{\includegraphics[width=1in]{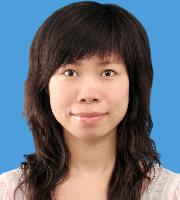}}]{Yufeng Chen}
received the B.S. degree from Beijing Jiaotong University, Beijing, China, in 2003, and the Ph.D. degree in pattern recognition and intelligent system from the Institution of Automation, Chinese Academy of Sciences, Beijing, in 2008. She joined Beijing Jiaotong University, Beijing, China, in 2014, as an Associate Professor. Her interests include natural language processing and machine translation.
\end{IEEEbiography}

\begin{IEEEbiography}[{\includegraphics[width=1in]{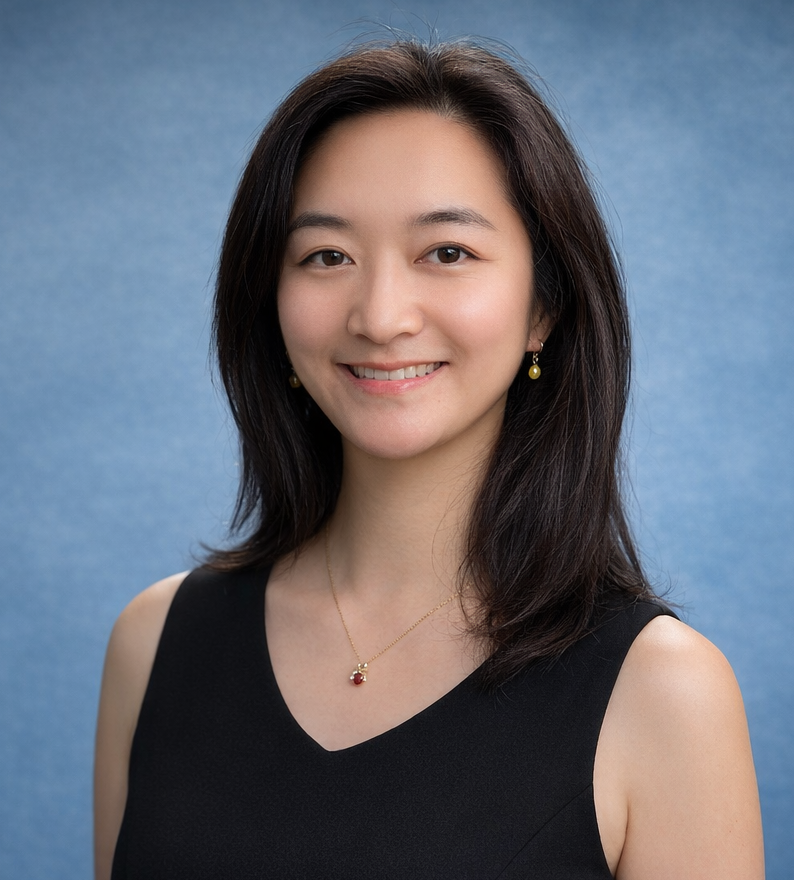}}]{Yang Zhao}
is a Lecturer (Assistant Professor) at the Department of Computer Science and Information Technology, La Trobe University. Previously, she was a Machine Learning Research Fellow at the Australian Institute for Machine Learning, The University of Adelaide. Her research interests lie broadly in the field of computer vision and deep learning, landmark detection and generative 3D modeling, AI for agriculture and AI for health.
\end{IEEEbiography}

\begin{IEEEbiography}[{\includegraphics[width=1in,height=1.25in,clip,keepaspectratio]{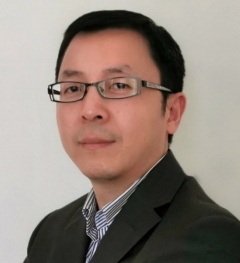}}]{Ling Shao} (Fellow, IEEE) is a Distinguished Professor with the UCAS-Terminus AI Lab, University of Chinese Academy of Sciences, Beijing, China. He was the founder of the Inception Institute of Artificial Intelligence (IIAI) and the Mohamed bin Zayed University of Artificial Intelligence (MBZUAI), Abu Dhabi, UAE. His research interests include generative AI, vision and language, and AI for healthcare. He is a fellow of the IEEE, the IAPR, the BCS and the IET.
\end{IEEEbiography}

\begin{IEEEbiography}[{\includegraphics[width=1in,height=1.25in,clip,keepaspectratio]{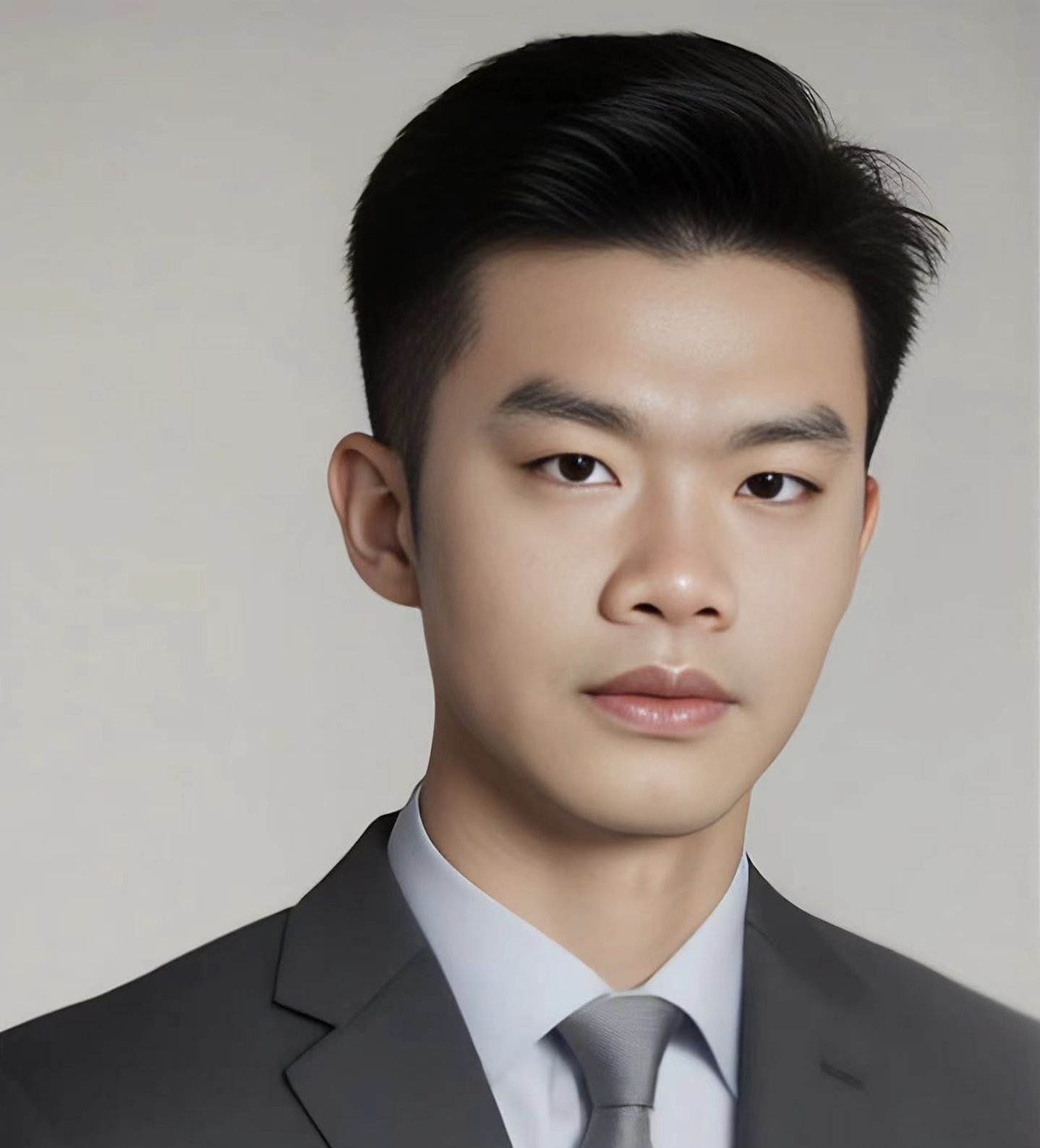}}]{Hao Tang}
is an Assistant Professor at Peking University, China. Previously, he held postdoctoral positions at CMU, USA, and ETH Zürich, Switzerland. He earned his master’s degree from Peking University, and his Ph.D. from the University of Trento, Italy.
He has had the opportunity to visit the University of Oxford, Northeastern University, NUS, and IIAI, among other institutions.
His research interests include computer vision, embodied AI, and generative AI, as well as their applications in scientific domains.
\end{IEEEbiography}

\end{document}